\documentclass{article}
\usepackage[final,nonatbib]{neurips_2023_ml4ps}
\usepackage[utf8]{inputenc} 
\usepackage[T1]{fontenc}    
\usepackage{hyperref}       
\usepackage{url}            
\usepackage{booktabs}       
\usepackage{amsfonts}       
\usepackage{nicefrac}       
\usepackage{microtype}      
\usepackage{xcolor}         
\usepackage{amsmath,amssymb}
\usepackage{amsthm}
\usepackage{multirow}
\usepackage{graphicx}
\usepackage{varwidth}

\usepackage{wrapfig}

\usepackage{algorithm}[compatible]
\usepackage{algpseudocode}
\usepackage{float}
\usepackage{enumitem}

\title{Extending Explainable Boosting Machines to \\ 
Scientific Image Data}
\author{%
  Daniel Schug\thanks{Equal contribution, ordered alphabetically} \\
  University of Maryland \\
  \texttt{dschug1@umd.edu} \\
  \And
  Sai Yerramreddy$^{*}$ \\
  University of Maryland \\
  \texttt{saiyr@umd.edu} \\
  \And
  Rich Caruana \\
  Microsoft Research \\
  \texttt{rcaruana@microsoft.com} \\
  \And
  Craig Greenberg \\
  National Institute of \\Standards and Technology \\
  \texttt{craig.greenberg@nist.gov} \\
  \And
  Justyna P. Zwolak\thanks{Author to whom any correspondence should be addressed.} \\
  National Institute of \\Standards and Technology \\
  \texttt{jpzwolak@nist.com} \\
}

\begin{document}
\maketitle
\begin{abstract}
As the deployment of computer vision technology becomes increasingly common in science, the need for explanations of the system and its output has become a focus of great concern.
Driven by the pressing need for interpretable models in science, we propose the use of Explainable Boosting Machines (EBMs) for scientific image data.
Inspired by an important application underpinning the development of quantum technologies, we apply EBMs to cold-atom soliton image data tabularized using Gabor Wavelet Transform-based techniques that preserve the spatial structure of the data.
In doing so, we demonstrate the use of EBMs for image data for the first time and show that our approach provides explanations that are consistent with human intuition about the data.
\end{abstract}

\section{Introduction}
\label{sec:intro}
Machine learning (ML)-based image analysis has found many applications throughout science, including analysis of data in particle physics~\cite{Aurisano_2016, Acciarri_2017}, dark matter searches~\cite{Alexander_2020, Khosa_2020} and quantum dots experiments~\cite{Zwolak21-AAQ, Usman2020}; predicting properties of materials~\cite{Kalidindi20, PhysRevMaterials.4.093801}; studying molecular representations and properties~\cite{duvenaud2015convolutional, butler2018machine}; and others in fields such as medicine and biology~\cite{caruana_intelligible, Letham_2015}. 
Given the widespread application of ML, there is a growing need for explainable ML to support applications that currently require human users to understand why a model provides the output it does. 

While the glass-box models, such as decision trees, linear regression, or classification rules, are relatively easy for humans to interpret, they tend to underperform when compared to the state-of-the-art black-box models such as deep neural networks (DNN). 
Moreover, glass-box models are not always easily adaptable to image data. 
At the same time, many experiments in the sciences produce data in the form of images, the nuanced analysis of which is limited by our preconceptions of the patterns and anomalies that could be present in the data.
While some of the analysis tasks, such as data pre-processing, classification, and feature detection, have been automated using black-box ML techniques, the complex relationship between inputs and outputs in such models makes them difficult to interpret.
This limits their application in areas where human-user understanding of the model output or the correlations between features implicitly utilized by the model are strictly necessary.
Prior work \cite{NEURIPS2018_294a8ed2} has demonstrated that some saliency methods, such as SmoothGrad~\cite{smilkov2017smoothgrad} or Gradient-weighted Class Activation Mapping (Grad-CAM)~\cite{Selvaraju_2019}, are independent of the model being explained as well as the data-generating process.  
This fundamental shortcoming calls into question whether these methods provide explanatory information or if they are more akin to edge detectors.

\begin{wrapfigure}[9]{R}{0.6\textwidth}
\vspace{-15pt}
  \begin{minipage}[c]{0.6\textwidth}
    \includegraphics[width=\textwidth]{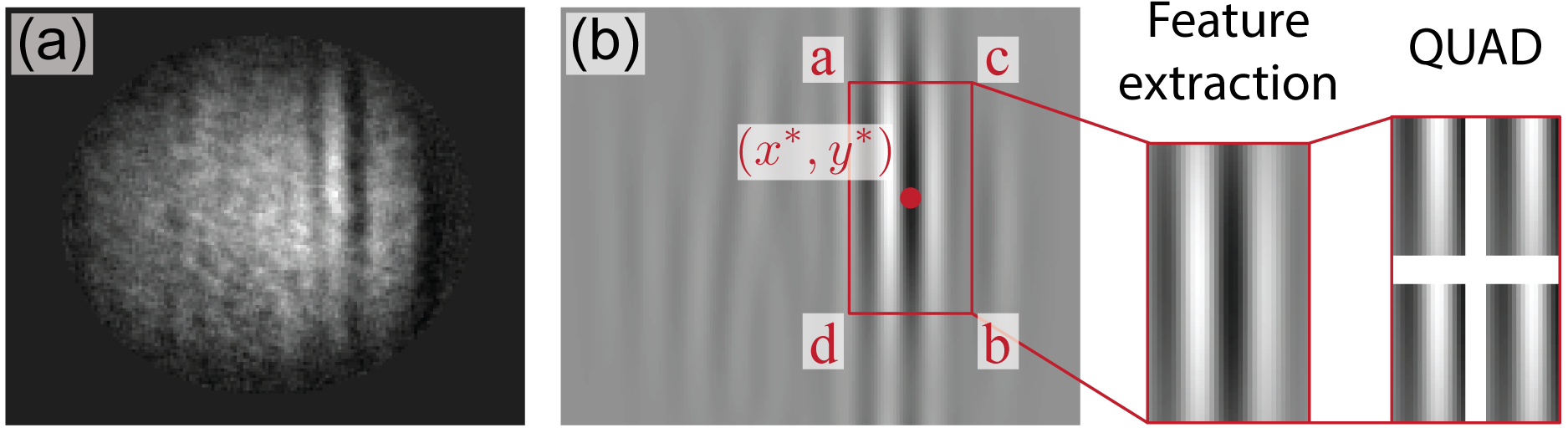}
  \end{minipage} \\
  \begin{minipage}[c]{0.6\textwidth}
    \caption{(a) Sample image from the reduced soliton dataset. 
  (b) The optimized Gabor filter for the image shown in (a).} 
  \label{fig:GWT}
  \end{minipage}
\end{wrapfigure}

In this work, we tackle the problem of ML explainability for image data using Explainable Boosting Machines (EBMs).
EBMs are models designed to be highly intelligible and explainable, while also achieving comparable accuracy against state-of-the-art ML methods~\cite{lou2013accurate}. 
However, to date, EBMs have only been used with tabular data and have not been adapted to any other data type.
An example of an image dataset where understanding the relative importance of and correlations between visual features characterizing the physical systems allows to differentiate between different states of the system is the {\it Dark solitons in Bose-Einstein condensates (BECs) dataset v.2.0}~\cite{SolitonDataset}. 
Solitonic excitations -- solitary waves that retain their size and shape and often propagate at a constant speed -- are present in systems at scales ranging from microscopic~\cite{Burger1999, Denschlag97}, to terrestrial~\cite{Russel1837, Hasegawa1973, Osborne1980, PhysRevLett.45.1095, Yomosa1984, Hashizume1985, Lakshmanan2009Tsunamis} and even astronomical~\cite{Stasiewicz2003}.
In images of BECs, solitons manifest as reductions in the atomic density (a ``dip'') surrounded by two ``shoulders'' of higher density.
While ideal solitons are vertically symmetric, many candidate excitations created experimentally break this symmetry, resulting in a number of physically-motivated subcategories, such as solitonic vortices or partial excitations~\cite{Mateo15-SDS}.

We focus on the subset of the soliton dataset consisting of data containing exactly one excitation [class-1 data, see Fig.~\ref{fig:GWT}(a)].
Each image in class-1 of the soliton dataset is tagged with the physically-motivated subcategory label and associated physics-based fits of an inverted and skewed Mexican-hat function to one-dimensional (1D) background-subtracted projections of solitonic excitations candidates obtained using a least-squares fit~\cite{Guo22-CMP}.
The original class-1 dataset is highly imbalanced, with the largest {\it longitudinal} class consisting of 2,229 images and the smallest, {\it clockwise vortices} class consisting of only 28 images.
We employ several strategies to mitigate this imbalance. 
Since feature orientation is important to our tabularization method, data labeled as {\it canted} is excluded from analysis due to the orientation variability combined with the small class size. 
The remaining 5-class dataset is further simplified into a 3-class dataset by combining data from the physically symmetric {\it top} and {\it bottom} partial soliton classes into a single category {\it partial} and the {\it clockwise} and {\it counterclockwise} solitonic vortices into a single category {\it vortex}. 
The data from the bottom partial and counterclockwise solitonic vortex classes are augmented via horizontal flipping to be consistent with the top partial and clockwise solitonic vortex classes, respectively.
We call the final dataset used in the experiments the {\it reduced soliton dataset}. The class split for the final 3 classes is as follows: 2,229 for the {\it longitudinal} class, 796 for the {\it partial} class, and 66 for the {\it vortex} class.

In this work, we investigate the utility of EBMs in classifying the reduced soliton data and compare their accuracy and explainability to other well-known methods. 
The contributions lie in three areas:

\begin{itemize}[topsep=-5pt,itemsep=0pt]
    \item We propose a novel Gabor wavelet transform (GWT)-based method to extract and tabularize visual features from images, moving beyond simpler, pixel-based approaches.
    \item We report a first-ever demonstration of applying EBMs to image data, verifying that EBMs can effectively reveal physically-meaningful patterns among the tabularized features.
    \item We show that our proposed approach provides better explanations than other state-of-the-art ML explainability methods for images.
\end{itemize}

\section{Background and Methods}
\label{sec:background}
This paper focuses on investigating whether EBMs can be adapted to classify image data.
EBMs are generalized additive models (GAMs) that account for pairwise interactions: 
    $g(E[y]) = \beta + \sum\nolimits_i f_i (x_i) + \sum\nolimits_{i \neq j} f_{ij}(x_i,x_j)$,
where $g$ is the link function~\cite{caruana_intelligible, lou2013accurate}. 
The contribution of each feature to a final prediction can be visualized and understood either through a plot of $f_i(x_i)$ vs. $x_i$ for the univariate terms or through heatmaps of the $f_{ij}(x_i,x_j)$ in the two-dimensional $x_i,x_j$-plane. 
This ability to analyze features either independently or, for the strongly interacting ones, as pairs is what makes EBMs highly intelligible. 
Moreover, since EBMs are additive models, they are modular in the sense that it is easy to reason about the contribution of each feature to the prediction~\cite{caruana_intelligible}.

To use EBMs, we need to automatically and in an interpretable way translate the primary visual features captured in images into a tabular representation.  
We do this using GWT and parameter space optimization. 
The GWT is a multi-scale multi-directional wavelet initially formulated for signal representation that currently sees modern use in feature detection in computer vision and image processing. 
It has been successfully used in computer vision tasks such as character recognition~\cite{wang_ding_liu_2005}, face detection~\cite{abhishree_latha_manikantan_ramachandran_2015} and iris detection~\cite{Daugman_2004}.
The GWT relies on convolving the Gabor kernel with an input image across a range of parameters to extract the characteristic features in the image.

The 2D GWT kernel is defined by the product of a Gaussian and a plane wave: 
\begin{equation*}
G_{\sigma_x,\sigma_y,\theta,\lambda}(x,y) = \nicefrac{1}{(\sqrt{2\pi}\sigma_x\sigma_y)}\, \exp(-(x^2/2\sigma_x^2 + y^2/2\sigma_y^2)\cdot \exp[i\cdot\lambda(x\cdot sin(\theta) + y\cdot cos(\theta))],
\end{equation*} 
where $\sigma_x$ and $\sigma_y$ represent scale in $x$ and $y$, and $\theta$ and $\lambda$ are the wave direction and wavelength.
By optimizing the parameters defining the Gabor kernel on a per-image basis, we can determine the \linebreak 

\begin{wrapfigure}[18]{R}{0.5\textwidth}
\vspace{-20pt}
\begin{minipage}{0.5\textwidth}
\begin{algorithm}[H]
\small
\caption{Gabor Quad Transform}
\label{alg:image_tabularization}
{\it {\bf Step 1.} Find $\sigma_x^*,\sigma_y^*,\lambda^*$ given image $u(x,y)$. }
\begin{algorithmic}[1]
\State {\bf Input:} image $u$, parameter spaces $\Sigma_x,\Sigma_y,\Lambda$
\State $\sigma_x^*,\sigma_y^*,\lambda^*=\mathrm{argmax}(||G_{\sigma_x,\sigma_y,\lambda}*u||_2)$ for $\sigma_x\in\Sigma_x,\sigma_y\in\Sigma_y,\lambda\in\Lambda$

\State {\bf Return:} optimal parameters ($\sigma_x^*,\sigma_y^*,\lambda^*$)
\end{algorithmic}
{\it {\bf Step 2.} Compute $\ell^2$ quad responses using optimal parameters from Step 1.}
\begin{algorithmic}[1]
\State {\bf Input:} image $u$, optimal parameters ($\sigma_x^*,\sigma_y^*,\lambda^*$)
\State with $u_{G^*}(x,y) = G_{\sigma_x^*,\sigma_y^*,\lambda^*}*u$ compute integral image $iu_{G^*} = \sum_{x'\leq x,y'\leq y} u_{G^*}(x,y)^2$
\State compute $x^*,y^*=\mathrm{argmax}((G_{\sigma_x,\sigma_y,\lambda}*u)(x,y)) $ 
\State compute quadrant responses $||u_{a,b,c,d}||_2=\sqrt{iu_{G^*}(a)+iu_{G^*}(d)-iu_{G^*}(b)-iu_{G^*}(c)}$ for points $a,b,c,d\in(x^*-\sigma_x^*,x^*,x^*+\sigma_x^*)\times(y^*-\sigma_y^*,y^*,y^*+\sigma_y^*)$
\State {\bf Return:} $\sigma_x^*,\sigma_y^*,\lambda^*,x^*,y^*,||u_{a,b,c,d}||_2$	
\end{algorithmic}
\end{algorithm}
\end{minipage}
\end{wrapfigure}

\vspace{-15pt}
locations and the size (scale) of the regions of interest (ROI). 
This is similar to the techniques used for blob-detection in scale-space theory~\cite{lindeberg_1998}, where features represent local maxima over the parameter space. 
The optimized kernels are then used to extract the local characteristics of those regions.
Finally, to translate those local characteristics into a tabular representation, the response of the data to the filter can be quantified using a histogram, the $\ell^2$-norm, the total variation norm, or other metrics.  
The procedural steps for the proposed tabularization algorithm in the form of pseudocode are presented in Algorithm~\ref{alg:image_tabularization}. 
The resulting tables are then used to train EBMs and their accuracy and explainability are compared against different convolutional neural networks (CNNs).

\section{Results}
\label{sec:experiment}
For the reduced solitonic excitation dataset, there is a single dominant ROI (at a single orientation) that corresponds to the primary excitation and its shoulders. 
Due to the unique expressiveness of the GWT, we can locate this region with a single filter.
This process consists of optimizing over the space of GWT parameters $\sigma_x\in\Sigma_x$, $\sigma_y\in\Sigma_y$, $\lambda\in\Lambda$, with $\theta = 0$. Specifically, $\sigma_x^*$ and $\sigma_y^*$ correspond to the optimized width and height of the excitation while $\lambda^*$ determines its optimized scale. 
In our implementation, it is beneficial to use knowledge of the image structure to aid in optimization, which we present in Algorithm~\ref{alg:param_optimization}. 
While computationally intensive compared with using a single predefined filter, this enables a reliable and accurate description of the excitation with features that completely describe the ROI for tabularization. 

\begin{wrapfigure}[9]{R}{0.38\textwidth}
\vspace{-35pt}
\begin{minipage}{0.38\textwidth}
\begin{algorithm}[H]
\small
\caption{Parameter optimization}
\label{alg:param_optimization}
{\it {\bf Step 1a.} Find $\sigma_y^*,\lambda^*$ given image $u(x,y)$. }
\begin{algorithmic}[1]
\State {\bf Input:} image $u$, large $\sigma_x^N$
\State $\sigma_y^*,\lambda^*=\mathrm{argmax}(||G_{\sigma_x^N,\sigma_y,\lambda}*u||_{2})$ 
\State {\bf Return:} optimal $(\sigma_y^*,\lambda^*)$
\end{algorithmic}
{\it {\bf Step 1b.} Find $\sigma_x^*$ given image $u(x,y)$. }
\begin{algorithmic}[1]
\State {\bf Input:} image $u$, large $\sigma_x^N$
\State $\sigma_x^*=\mathrm{argmax}(||G_{\sigma_x,\sigma_y^*,\lambda^*}*u||_{2})$ 
\State {\bf Return:} optimal $\sigma_x^*$
\end{algorithmic}
\end{algorithm}
\end{minipage}
\end{wrapfigure}

Given the role of symmetry in distinguishing different classes of excitations, our algorithm extracts the magnitudes in each quadrant of the region obtained by the optimized parameters. 
These subregions are found by describing the region around $(x^*,y^*) = \mathrm{argmax}((G_{\sigma_x^*,\sigma_y^*,\lambda^*}*u)(x,y))$ bounded by $(\pm\sigma_x^*,\pm\sigma_y^*)$. 
Each subregion is measured in terms of its $\ell^2$-norm to acquire a single numerical description of the response. 
For each image, this produces raw features of the form $||(G*u)^{>x^*}_{>y^*}||_2$, $||(G*u)^{>x^*}_{<y^*}||_2$, $||(G*u)^{<x^*}_{>y^*}||_2$, $||(G*u)^{<x^*}_{<y^*}||_2$, 
where $||(G*u)^{(\cdot)}_{(\cdot)}||_2$ is the corresponding integral image in the respective quadrants.

\textbf{Classification Experiments.}
We assess the classification performance of various methods using accuracy, precision, and recall. 
We carry out five 6-fold stratified cross-validations and report averaged results, with precision and recall analyzed at the class level.
The experiments involve benchmarking EBMs on the physics-based fits (PF+EBM) and testing new methods using tabularized features from GWT (GF) and EBMs ({\it GF+EBM}). 
We also use EBMs on data representing GWT features and the physics-based fits ({\it GF+PF+EBM}).
All tests using EBMs are performed in the one-versus-rest fashion.

For comparison with more advanced classification techniques, we test several NN-based classification methods.
Due to the challenges posed by the soliton dataset, including small dataset size, grayscale images, and class imbalance, we opt for smaller NN models.
We train a  DNN with just fully connected layers, a fine-tuned convolutional autoencoder (CAE+NN), and a simple CNN.
All NN models were trained using the Adam optimizer~\cite{kingma2017adam} with a scheduled learning rate and early stopping callback.

\begin{table}[t]
\small
\renewcommand{\arraystretch}{1.0}
\renewcommand{\tabcolsep}{2.2pt}
  \caption{The table presents the results of 6-fold cross-validation for the reduced soliton dataset.
  The results are sorted by method (columns) and by data class (rows).}
  \label{tab:3-class-table}
  \centering
  \begin{tabular}{llccccccccc}
    \toprule
    \multirow{2}{*}{Class} & \multirow{2}{*}{Metric} & PF & GF & GF+PF & \multirow{2}{*}{DNN} & \multirow{2}{*}{CNN} & \multirow{2}{*}{CAE+NN} & EGF & EGF+PF \\
    \cmidrule{3-5} \cmidrule{9-10}
    & & \multicolumn{3}{c}{+EBM} & & & & \multicolumn{2}{c}{+EBM} \\
    \cmidrule{1-10}
    All & Accuracy & 87.4(4) & 84.8(6) & {\bf 91.4(4)} & 82.3(8) & 86.9(9) & {\bf 91.9(5)} & 88.0(4) & {\bf 92.8(5)}\\
    \midrule
    \multirow{2}{*}{Longitud.} & Precision & 89.4(2) & 87.4(6) & 92.7(6) & 88.2(1.3) & 88.8(7) & {\bf 95.4(8)} & 90.4(3) & {\bf 94.0(4)} \\
    & Recall & 95.7(4) & 93.9(7) & {\bf 96.9(4)} & 93.0(1.3) & 95.8(1.4) & 94.3(5) & 95.1(5) & {\bf 97.4(2)}\\
    \midrule
    \multirow{2}{*}{Partial} & Precision & 81.0(1.2) & 76.3(1.8) & {\bf 87.5(1.1)} & 83.0(3.5) & 84.6(2.5) & 85.4(1.9) & 87.7(7) & {\bf 89.1(1.2)}\\
    & Recall & 71.4(7) & 66.2(2.1) & 81.0(2.0) & 64.6(3.2) & 67.9(3.6) & {\bf 84.5(2.7)} & 73.3(8) & {\bf 83.7(1.4)}\\
    \midrule
    \multirow{2}{*}{Vortex} & Precision & 22.6(20.9) & 16.7(11.5) & {\bf 83.9(4.4)} & 17.7(10.1) & 26.8(8.8) & 62.3(12.4) & 67.3(4.6) & {\bf 87.5(2.9)}\\
    & Recall & 1.8(1.8) & 1.5(1.0) & 31.8(3.3) & 50.7(9.4) & 43.8(6.4) & {\bf 96.6(2.8)} & 22.4(8.5) &  44.5(3.1)\\
    \bottomrule
  \end{tabular}
  \vspace{-15pt}
\end{table}

The results from all experiments are presented in Table~\ref{tab:3-class-table}. 
The {\it GF+PF+EBM} method is on par with the {\it CAE+NN} method in terms of accuracy, with $91.4(4)~\%$ and $91.9(5)~\%$, respectively.
On a per-class level, we observe that the {\it GF+PF+EBM} method has comparable precision and recall to the {\it CAE+NN} method for the longitudinal and partial classes. 
For the vortex class, the {\it GF+PF+EBM} method has significantly better precision, whereas {\it CAE+NN} method has a better recall. 

\begin{figure}[t]
  \centering
  \includegraphics[width=\linewidth]{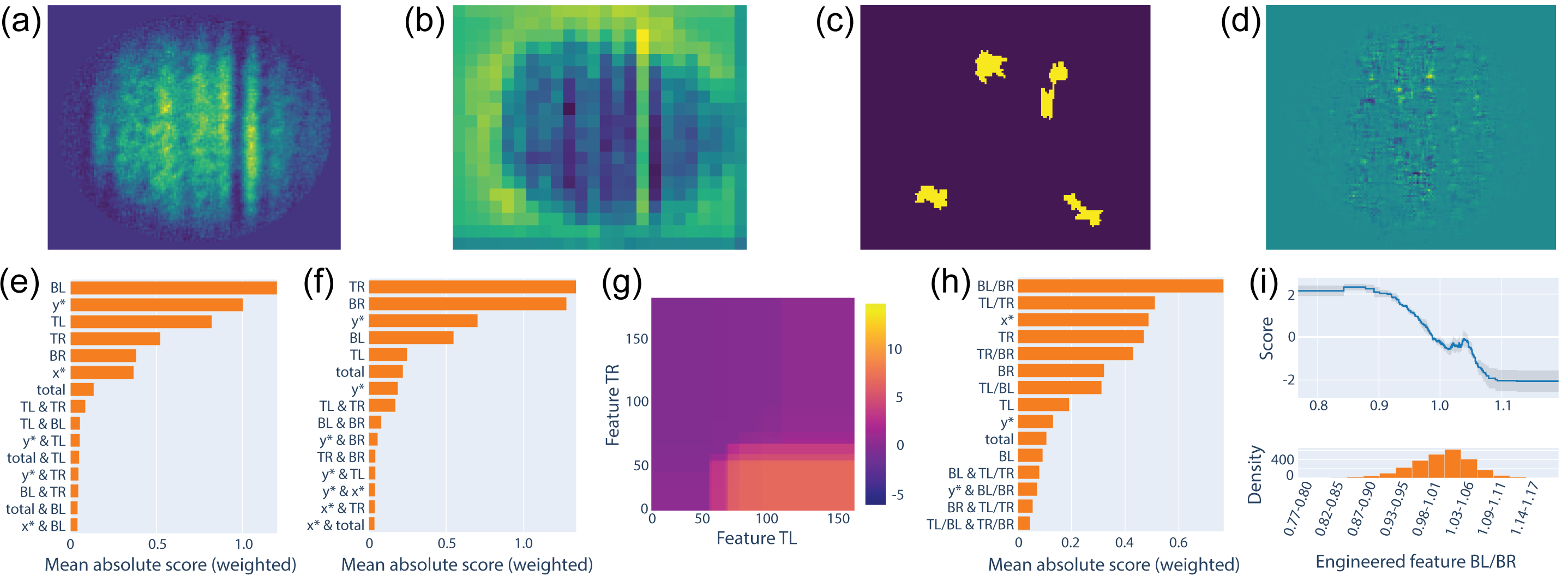}
  \caption{(a) Sample image from the reduced soliton dataset.
  (b)-(d) Interpretability result for the CAE+NN method: (b) Grad-CAM heatmap, (c) LIME mask, (d) SHAP heatmap. 
  (e)-(i) Interpretability results for the EBMs: feature importance for the (e) partial and (f) vortex classes, (g) pairwise interaction for TL and TR features for the vortex class, (h) engineered features importance for vortex class, and (i) the BL/BR engineered feature dependence plot.}
  \label{fig:EBM_models}
\end{figure}

\textbf{Interpretability Results.}
From the interpretability results for the {\it CAE+NN} approach in Figs.~\ref{fig:EBM_models}(b)-(d), we can see that the Grad-CAM approach tends to take relevant parts of the input into account when producing the output.
For LIME, the overlap between the relevant input region and the mask is small. 
However, LIME has trouble dealing directly with grayscale data and requires an intermediate conversion from grayscale to RGB which likely affects its performance \footnote[1]{As noted by the authors of the LIME open-source package \cite{limegithubissues}}. 
SHAP seems to highlight regions around the relevant area as well, although it additionally highlights a lot of the non-relevant regions. 
While interpretability approaches such as LIME, GradCAM, and SHAP offer valuable insights, they have inherent limitations that we seek to overcome. 
First, both LIME and GradCAM are very sensitive to perturbations, and using a smoothing technique is not always viable due to the nature of the image data \cite{smilkov2017smoothgrad}.
LIME and SHAP have parameters that need to be optimized, such as surrogate model type, superpixel segmentation algorithm and distance metric for LIME, and estimation method and number of samples for SHAP.
Choosing optimal parameters is time-consuming while sub-optimal parameters can significantly impact the quality of the explanations produced and slow down the analysis.
All three methods can struggle to capture complex interactions between features, which can lead to inaccurate or incomplete explanations.
Finally, the NN-based approaches are not user-friendly as localizing and analyzing the region of interest programmatically is difficult and would require a lot of additional computation. We aim to address these limitations using our {\it EGF+EBM} approach.
 
In the first series of experiments with EBMs, we test the EBM on the raw GF consisting of top left (TL), top right (TR), bottom left (BL), and bottom right (BR) measures as well as the optimized excitation center $(x^*,y^*$). 
These measures correspond to the intensity of the GWT filter response in each quadrant of the extracted ROI shown in Fig.~\ref{fig:GWT}.
EBMs provide a global explanation for all the predictions by graphically depicting the contribution of individual features and pairwise correlations to the model. For example, Fig.~\ref{fig:EBM_models}(e) and Fig.~\ref{fig:EBM_models}(f) show feature importance for class partial and vortex, respectively.

Notably, for the partial class, we observe that the offset in the $y$ direction ($y^*$) emerged as one of the most significant features, aligning with human intuition due to the nature of partial excitations occurring in the upper half of the BEC. 
In contrast, for the vortex class, the expected asymmetry with respect to the $x$-axis is confirmed by EBMs consistently ranking the TR and BR quadrants as the most important. 
The pairwise interaction map for these two features shown in Fig.~\ref{fig:EBM_models}(g) further confirms this intuitive dependence. 
Such correlations can guide the creation of new features capturing such interactions, potentially enhancing model performance. 
Indeed, adding engineered GF (EGF) features representing primary pairs (TL/BL, TR/BR, TL/TR, and BL/BR) resulted in substantial performance improvements, particularly for the underrepresented vortex class as seen in Table~\ref{tab:3-class-table}.

Figure~\ref{fig:EBM_models}(h) further confirms the importance of EGF features, with all four included in the list of top 7 features, which indicates that they indeed provide additional predictive value.
Figure~\ref{fig:EBM_models}(i) shows that $BL/BR<1$ (i.e., stronger response in BR than in BL quadrant) indicates a vortex characteristic, which, again, agrees with the intuition. 
Similar analysis can be carried out for the remaining features and for all classes further validating the interpretability of the FGs and EFGs.

\section{Conclusions and Future Work}
\label{sec:concl}
In this paper, we explore the application of GWT-based feature extraction and parameter optimization to extend EBMs for image data analysis. 
Comparative assessments against state-of-the-art DNN methods reveal that the GWT-based approach achieves comparable accuracy while also producing intuitive explanations. 
While our experiments focus on soliton in BEC data, our methods can readily adapt to other image-based scientific datasets with underlying visual structures. 
For more complex data requiring considerations of feature orientation or scale, the design of a tailored GWT-based filter bank may be necessary.
Future research directions include applying these techniques to diverse scientific datasets, such as additional cold atoms dataset and medical datasets. We also plan to investigate their utility in clustering unlabeled data with unknown class counts.

\begin{ack}
The views and conclusions contained in this paper are those of the authors and should not be interpreted as representing the official policies, either expressed or implied, of the U.S. Government. 
The U.S. Government is authorized to reproduce and distribute reprints for Government purposes notwithstanding any copyright noted herein. 
Any mention of equipment, instruments, software, or materials; it does not imply recommendation or endorsement by the National Institute of Standards and Technology.
\end{ack}


\end{document}